\documentclass[sigconf, nonacm]{acmart}

\usepackage{listings}
\usepackage{titlesec}
\usepackage{balance}
\usepackage{enumitem}

\newcommand{\fig}[1]{Fig.~\ref{#1}}
\newcommand{\sect}[1]{\S\ref{#1}}

\newcommand{\eg}{\textit{e.g.}, }

\newcommand{\etal}{\textit{et al.}, }

\titlespacing{\section}{0pt}{1.0em}{0.3em}
\titlespacing{\subsection}{0pt}{0.5em}{0.3em}
\titlespacing{\subsubsection}{0pt}{0.5em}{0.3em}
\titleformat*{\subsection}{\normalsize\bfseries}
\titleformat*{\section}{\Large\bfseries}

\usepackage{color}

\definecolor{dkgreen}{rgb}{0,0.6,0}
\definecolor{gray}{rgb}{0.5,0.5,0.5}
\definecolor{mauve}{rgb}{0.58,0,0.82}
\newcommand\footnotewithoutmarker[1]{%
  \begingroup
  \renewcommand\thefootnote{}\footnote{#1}%
  \addtocounter{footnote}{-1}%
  \endgroup
}

\begin{document}
\title[End-to-end Text-to-SQL Generation within an Analytics Insight Engine]{End-to-end Text-to-SQL Generation within an\\ Analytics Insight Engine}

\author{Karime Maamari}
\affiliation{%
  \institution{Distyl AI}
}
\email{karime@distyl.ai}

\author{Amine Mhedhbi}
\affiliation{%
  \institution{Polytechnique Montréal}
}
\email{amine.mhedhbi@polymtl.ca}

\begin{abstract}
Recent advancements in Text-to-SQL have pushed database management systems towards greater democratization of data access. 
Today's language models are at the core of these advancements. They enable  impressive Text-to-SQL generation as experienced in the development of Distyl AI's Analytics Insight Engine. 
Its early deployment with enterprise customers has highlighted three core challenges.  
First, data analysts expect support with authoring SQL queries of very high complexity. 
Second, requests are ad-hoc and, as such, require low latency. 
Finally, generation requires an understanding of domain-specific terminology and practices. 

The design and implementation of our Text-to-SQL generation pipeline, powered by large language models, tackles these challenges. 
The core tenants of our approach rely on external knowledge that we extract in a pre-processing phase, on retrieving the appropriate external knowledge at query generation time, and on decomposing SQL query generation following a hierarchical CTE-based structure.
Finally, an adaptation framework leverages feedback to update the external knowledge, in turn improving query generation over time.  
We give an overview of our end-to-end approach and highlight the operators generating SQL during inference. 

\end{abstract}

\maketitle

\footnotewithoutmarker{$^{*}$Compound AI Systems Workshop at AI + Data Summit 2024}
\vspace*{-22pt}
\section{Introduction}

Distyl AI's Analytics Insight Engine supports data analysts in ideation, query authoring, and exploratory data analysis. 
At the core of this engine is a Text-to-SQL module powered by large language models (LLMs). 
To provide a high-quality user experience across these tasks, 
we identified four Text-to-SQL specifications:\\

\noindent\textbf{\emph{• Generalization:}} Ability to easily bootstrap a pipeline within a new environment to produce complex queries. 
Appendix~(\sect{sec:generation-example}) shows an example of  such a query generated by the engine.\\
\textbf{\emph{• Latency:}} Ability to generate SQL in at most 60 seconds and on average in 30 seconds.\\
\textbf{\emph{• Robustness:}} Ability to maintain the same generation quality across multiple runs of the same input query. \\
\textbf{\emph{• Improvement:}} Ability to dynamically improve through feedback from users or query execution.\\

The flexibility of LLMs makes meeting these specifications feasible. Instead of rule-based techniques that invoke task-specific models~\citep{xu2017sqlnet, zhong2017seq2sql, dong2018coarsetofine, sun2018semantic, dialsql}, 
the focus now is on external knowledge management, prompting, and fine-tuning for generation~\cite{pourreza2023dinsql, gao2023texttosql, wang2024macsql, talaei2024chess, lee2024mcssql}. 
The latter approach leads to state-of-the-art results and is the one we adopt ~\cite{talaei2024chess, lee2024mcssql, yu2019spider, li2023llm}. 

We rely on pre-processing to construct an external knoweledge set. At inference, we use a subset of relevant knowledge for query generation using multiple LLM calls. Finally, we rely on adaptive learning from feedback to update the knowledge set, as described briefly below:\\

\begin{figure*}[t]
    \centering
    \captionsetup{justification=centering}
    \includegraphics[scale=0.85]{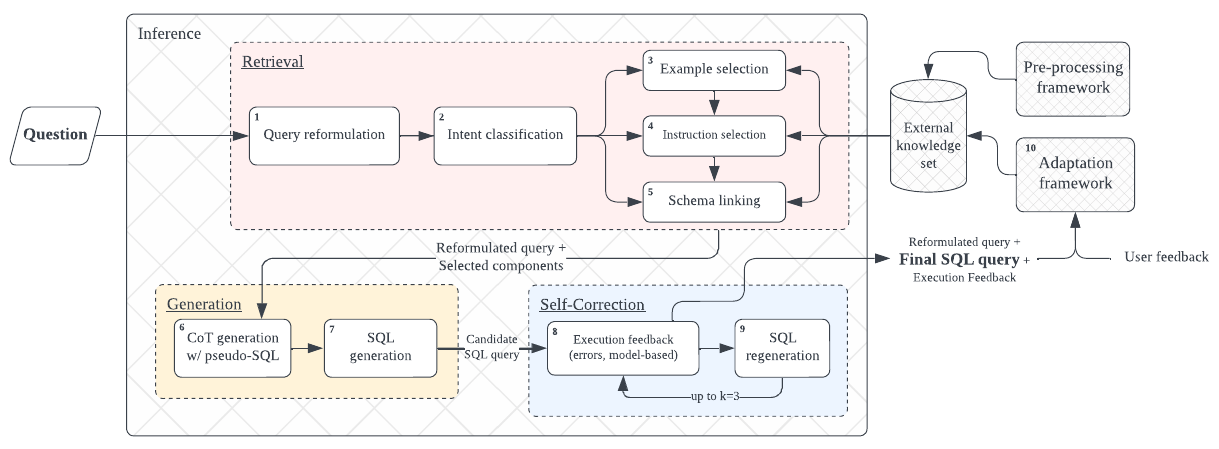}
    \vspace{-1.2em}
    \caption{Overview of Text-to-SQL generation pipeline with 3 steps: i) external knowledge retrieval to add context; ii) LLM SQL generation task; and iii) self-correction, which re-generates based on syntactic and semantic feedback from execution.}
    \label{fig:flow}
\end{figure*}

\noindent\textbf{\emph{• Pre-processing}}\textbf{:}  
This phase takes as input i) SQL queries from logs of prior executions and ii) documents that contain domain-specific terminology and practices. As output, it produces an external knowledge set that contains examples and instructions, both partitioned by user intent, as well as representation of the schema. We explain further each of these components in Section~\sect{sec:pre-processing}.\\
\noindent\textbf{\emph{• Inference}}\textbf{:} Given a natural language input query, a pipeline made of multiple operators generates an output SQL query. The generation is done using LLM calls. 
\fig{fig:flow} shows the inference pipeline and its operators.\\ 
\noindent\textbf{\emph{• Adaptation}}\textbf{:} Based on feedback from users, as well as syntactic and semantic execution errors, this phase updates the external knowledge set to improve future SQL generations. 

\section{Pre-processing Phase}
\label{sec:pre-processing}
In pre-processing, we construct a set of examples, instructions, and a database schema using
query logs and domain-specific documents. 
Below, we describe the representation of each component. 
Appendix \sect{sec:appendix-representation} shows the structure of the representations. 

\subsection{Example Representation}
Our examples are SQL queries are represented as decomposed sub-expressions, with an associated natural language query. This is in contrast with traditional examples where the SQL queries are represented in their raw format. The decomposed sub-expressions, decompositions for short, are obtained as follows. 

We take as input full SQL queries from historical logs or directly from domain experts. 
We first reformat them into a CTE-based sketch. 
Then, each reformatted example is decomposed into sub-queries based on WITH clauses, and finally into sub-expressions based on inner clauses. 
As a result, we represent each query example in a hierarchically decomposed form (shown in Appendix~\sect{sec:A1}). 
We augment this decomposed form with natural language descriptions of the complex operations, tables, and sub-queries involved. 
Our CTE-based sketch and associated decomposition is a novel approach, though it relies upon the insights of prior sketch-based slot-filling \citep{xu2017sqlnet} and decomposed generation approaches \citep{wang2024macsql, DBLP:conf/nips/PourrezaR23}. 

\subsection{Instruction Representation}
Generation instructions are guidelines provided to the model, detailing how to interpret the natural language input and formulate it into a correct SQL query. 
We obtain the instructions from example queries and documents containing domain-specific terminology and practices. We represent them in natural language, with expected SQL sub-expressions when relevant (shown in Appendix~\sect{sec:A2}).

\subsection{Schema Representation}
The schema representation is an outline of the database's structure which informs the model about the elements available for constructing SQL queries. 
We obtain the schema directly from the database or from database documentation, and represent it with table names, column names, column types, column descriptions, and column data samples (shown in Appendix~\ref{sec:A3}). 

\section{Inference Phase}
\label{sec:inference}
Given an input query, we retrieve relevant external knowledge, then we generate a candidate SQL query and apply self-correction if necessary.\\

\vspace*{-1.25em}
\subsection{External Knowledge Retrieval}

Before retrieval, we reformulate the input user query in a chosen canonical form and identify the user intent. 
We then retrieve, in order, relevant examples, instructions, and schema elements. We use the results of each retrieval for the that of the subsequent ones, \eg we use retrieved examples when retrieving instructions. 

First, we use the user intent to retrieve an example subset. Recall that examples are partitioned by intent in the pre-processing phase. We then re-rank the examples based on similarity to the reformulated query. 
Second, we retrieve instructions in a similar fashion to examples, except that we re-rank using, in addition, the selected examples. 
Finally, for the schema, we use LLM calls to identify irrelevant elements, as done by Wang \etal\cite{wang2024macsql}.\\
\vspace*{-1.25em}
\subsection{Generation}

We generate a candidate SQL query as follows. 
First, we construct a Chain-of-thought (CoT) reasoning plan~\cite{wei2023chainofthought, kojima2023large, wang2023planandsolve} from the reformulated input query and retrieved external knowledge. 
We then augment this reasoning plan with pseudo-SQL from selected examples.
Finally, we rely on an LLM call, using the reformulated query, retrieved external knowledge, and CoT plan to generate the candidate SQL query.
\vspace*{-1.25em}
\subsection{Self-Correction}

We now execute the generated candidate SQL query and, if necessary, apply corrections.
First, we obtain feedback from syntax-based errors and model-based assessments, which use predefined criteria to gauge the correctness of the output \citep{talaei2024chess}. 
We then correct the query by adding the feedback to a new LLM call for re-generation~\citep{wang2018robust}. 

\section{Related Work}

Early approaches struggled to incorporate external knowledge in a flexible manner due to contextual limitations~\citep{zhong2017seq2sql}. However, they could still extract relevant schema elements using semantic parsers that linked input queries with database schemas \citep{yu2018syntaxsqlnet}. Conversely, modern LLM-based approaches benefit from their ability to seamlessly integrate relevant knowledge at the query generation stage, leading to improved accuracy and reliability \citep{gao2023texttosql, lee2024mcssql}.

Query generation was heavily restricted by the use of query sketches in early approaches, limiting the potential output space of queries \citep{xu2017sqlnet}. Whereas, LLMs' flexibility allows modern solutions to explore a wider array of potentially more complex and varied SQL queries to generate~\citep{gao2023texttosql, talaei2024chess}.

Self-correction was initially introduced as execution-guided decoding, where queries were corrected based on execution results \citep{wang2018robust}. Current approaches extend this capability by not only responding to execution errors but also adapting queries based on a broader set of predefined criteria, enhancing the variety of potential corrections \citep{talaei2024chess}.

\section{Conclusion}
Our Text-to-SQL pipeline enables the production of complex SQL queries, taking into account automatic bootstrapping in new deployments through its pre-processing phase, as well as both user and system feedback through its adaptation phase. 
We believe that an empirical analysis of pipelines similar to the one we describe can lead to a better understanding of cross-cutting issues and provide practitioners with better guidance. % Furthermore, it can help move us towards a more dynamic approach to selecting parameters

\clearpage

\balance
\bibliographystyle{abbrv}
\bibliography{references}

\clearpage
\onecolumn
\appendix
\section{Generation Example}
\label{sec:generation-example}
\small
\subsection{Input - Prompt for Language Model}
\begin{verbatim}
### Input Query 

Identify the top 5 sports associations with the best and worst quarter-over-quarter financial performance in the United 
States for Q2 2023. 

### Schema Representation

- Table: SPORTS_FINANCIALS
  - Columns:
     - COUNTRY (text)
       - Description: The country where the financial data is applicable.
       - Sample rows: ['UNITED STATES', 'CANADA', 'UNITED KINGDOM']
     - SPORT_CATEGORY (text)
       - Description: The category of sport such as basketball, football, baseball, etc.
       - Sample rows: ['Basketball', 'Football', 'Baseball']
     - ...
- Table: SPORTS_VIEWERSHIP
  - Columns:
     - COUNTRY (text)
       - Description: The country where the viewership data is recorded.
       - Sample rows: ['UNITED STATES', 'CANADA', 'UNITED KINGDOM']
     - VIEW_MONTH (date)
       - Description: The month for which viewership hours are recorded.
       - Sample rows: ['2023-01-01', '2023-02-01', '2023-03-01']
     - ...
- Foreign keys: 
  - (SPORTS_FINANCIALS, SPORTS_VIEWERSHIP): 
    - (COUNTRY, COUNTRY)
  ...

### Intent-specific Instructions: 
   
1. Performance Change Calculation: Apply a -1 multiplier when calculating the change in performance metrics. 
   e.g. COST_CHANGE = -1 * (CURRENT_COST - PRIOR_COST)

...
   
10. Conditional aggregation: Use conditional aggregation when comparing revenue data across periods. 
    Employ the CASE statement within the aggregation function to isolate the value of interest for each period, 
    facilitating...

### Example Decompositions

{
    "input_nl": "Top 5 sports by change in RPV from 2021 to 2022",
    "complex_terms": ['RPV: RPV is defined as..., its SQL representation is...']
    ...
    "cte_1_columns": {
        "SELECTs/CALCs": [
            "COUNTRY",
            "SPORT_CATEGORY",
        ],
    ...
    "final_columns": {
        "SELECTs/CALCs": ["SPORT_RANK", "SPORT_CATEGORY", "RPV_CHANGE"],
        "WHEREs": ["SPORT_RANK <= 5"],
        "ORDERs": ["SPORT_RANK"]
    }
}
\end{verbatim}

\clearpage 
\onecolumn

\subsection{Output - SQL Generated by Language Model}
\label{subsec:output-sql-generated-by-language-model}

\begin{lstlisting}[language=SQL]
WITH FINANCIALS AS (
  SELECT
    COUNTRY,
    SPORT_CATEGORY,
    SUM(CASE WHEN TO_CHAR(FIN_MONTH, 'YYYY"Q"Q') = '2023Q1' THEN REVENUE ELSE 0 END) AS REVENUE_2023Q1,
    SUM(CASE WHEN TO_CHAR(FIN_MONTH, 'YYYY"Q"Q') = '2023Q2' THEN REVENUE ELSE 0 END) AS REVENUE_2023Q2
  FROM
    SPORTS_FINANCIALS
  WHERE
    TO_CHAR(FIN_MONTH, 'YYYY"Q"Q') IN ('2023Q1', '2023Q2')
    AND COUNTRY = 'UNITED STATES'
  GROUP BY
    COUNTRY,
    SPORT_CATEGORY
), VIEWERSHIP AS (
  SELECT
    COUNTRY,
    SUM(CASE WHEN TO_CHAR(VIEW_MONTH, 'YYYY"Q"Q') = '2023Q1' THEN VIEWER_HOURS ELSE 0 END) AS VIEWER_HOURS_2023Q1,
    SUM(CASE WHEN TO_CHAR(VIEW_MONTH, 'YYYY"Q"Q') = '2023Q2' THEN VIEWER_HOURS ELSE 0 END) AS VIEWER_HOURS_2023Q2
  FROM
    SPORTS_VIEWERSHIP
  WHERE
    TO_CHAR(VIEW_MONTH, 'YYYY"Q"Q') IN ('2023Q1', '2023Q2')
    AND COUNTRY = 'UNITED STATES'
  GROUP BY
    COUNTRY
), CALCULATIONS AS (
  SELECT
    f.SPORT_CATEGORY,
    CAST(f.REVENUE_2023Q2 AS FLOAT) / NULLIF(v.VIEWER_HOURS_2023Q2, 0) AS RPV,
    CAST(f.REVENUE_2023Q1 AS FLOAT) / NULLIF(v.VIEWER_HOURS_2023Q1, 0) AS PRIOR_QTR_RPV,
    -1 * ((CAST(f.REVENUE_2023Q2 AS FLOAT) / NULLIF(v.VIEWER_HOURS_2023Q2, 0)) - (CAST(f.REVENUE_2023Q1 AS FLOAT) / NULLIF(v.VIEWER_HOURS_2023Q1, 0))) AS RPV_CHANGE,
    (-1 * ((CAST(f.REVENUE_2023Q2 AS FLOAT) / NULLIF(v.VIEWER_HOURS_2023Q2, 0)) - (CAST(f.REVENUE_2023Q1 AS FLOAT) / NULLIF(v.VIEWER_HOURS_2023Q1, 0)))) * NULLIF(v.VIEWER_HOURS_2023Q2, 0) as IMPACT,
    ROW_NUMBER() OVER (PARTITION BY f.COUNTRY ORDER BY (-1 * ((CAST(f.REVENUE_2023Q2 AS FLOAT) / NULLIF(v.VIEWER_HOURS_2023Q2, 0)) - (CAST(f.REVENUE_2023Q1 AS FLOAT) / NULLIF(v.VIEWER_HOURS_2023Q1, 0)))) DESC) AS SPORT_RANK,
    ROW_NUMBER() OVER (PARTITION BY f.COUNTRY ORDER BY (-1 * ((CAST(f.REVENUE_2023Q2 AS FLOAT) / NULLIF(v.VIEWER_HOURS_2023Q2, 0)) - (CAST(f.REVENUE_2023Q1 AS FLOAT) / NULLIF(v.VIEWER_HOURS_2023Q1, 0)))) ASC) AS WORST_SPORT_RANK
  FROM
    FINANCIALS f
  JOIN
    VIEWERSHIP v ON f.COUNTRY = v.COUNTRY
)
SELECT
  SPORT_RANK, SPORT_CATEGORY, RPV, PRIOR_QTR_RPV, RPV_CHANGE, IMPACT
FROM
  CALCULATIONS
WHERE
  SPORT_RANK <= 5 OR WORST_SPORT_RANK <= 5
ORDER BY
  SPORT_RANK;
\end{lstlisting}

\normalsize
\section{External Knowledge Representation - After Pre-Processing}
\label{sec:appendix-representation}

\subsection{Examples}
\label{sec:A1}
\begin{verbatim}
"input_nl": <str: the input natural language query>,
"complex_terms": <list(str): complex calcs required>,
"features": {
    "tables": <list(str): tables used and>,
    "CTEs": <int: number of CTEs>,
    "CTE_desc": <list(str): brief descriptions of each CTE>
},
"cte_1_columns": {
    "SELECTs/CALCs": <list(str): columns and distinct calcs/aggs used for CTE 1>,
    "JOINs": <list(str): distinct join clauses in CTE 1>,
    "WHEREs": <list(str): distinct where clauses for CTE 1>,
    "GROUP_BYs": <list(str): columns used in group by clause for CTE 1>,
    "ORDERs": <list(str): columns used in order by clause in CTE 1>,
    "LIMITs": <list(str): limit clauses in CTE 1>
},
...
"cte_N_columns": {
    ...
},
"final_columns": {
    "SELECTs/CALCs": <list(str): columns and distinct calcs/aggs used for final>,
    "JOINs": <list(str): distinct join clauses in final>,
    "WHEREs": <list(str): distinct where clauses for final>,
    "GROUP_BYs": <list(str): columns used in group by clause for final>,
    "ORDERs": <list(str): columns used in order by clause in final>,
    "LIMITs": <list(str): limit clauses in final>
},
"full_sql_query": <str: the full SQL query> 

\end{verbatim}
\subsection{Instructions}
\label{sec:A2}
\begin{verbatim}
1. <str: instruction in natural language>
   <sql: representation, either default or related>
...
N. <str: instruction in natural language>
   <sql: representation, either default or related>
   
\end{verbatim}
\subsection{Schema}
\label{sec:A3}
\begin{verbatim}
- Table: <str: table name>
  - Columns:
     - <str: column name> (<str: column type>)
       - Description: <str: natural language description>
       - Sample rows: <list: top 5 most frequent, 
                             or all (if <= 10)>
     ...
  - Primary key: <str: column name>
...
- Foreign keys: 
  - (<str: table name 1>, <str: table name N>): 
    - (<str: key 1>, <str: key N>)
    ...
  ...
\end{verbatim}

\end{document}